\documentclass[letterpaper, 10 pt, conference]{ieeeconf} 
\usepackage{comment}
\usepackage{epsfig}
\usepackage{graphicx}
\usepackage{amsmath}
\usepackage{amssymb}
\usepackage{graphicx}
\usepackage{url}
\usepackage{array}
\usepackage{multirow}
\usepackage{color}
\usepackage{breqn}
\usepackage{fancyhdr}
\newcommand{\etal}{\textit{et al}. }
\usepackage{amsmath}
\usepackage[mathscr]{eucal}
\DeclareMathOperator*{\argmax}{argmax}
\usepackage{multicol}
\usepackage[noend]{algpseudocode}
\usepackage{algorithm}
\usepackage{subfigure}

\IEEEoverridecommandlockouts                              
\title{AutoFCL: Automatically Tuning Fully Connected Layers for Handling Small Dataset}

\author{S.H.Shabbeer Basha, Sravan Kumar Vinakota, Shiv Ram Dubey, Viswanath Pulabaigari, Snehasis Mukherjee \\
Indian Institute of Information Technology Sri City, India.
}

\usepackage{color}
\usepackage{fancyhdr}
\fancypagestyle{specialfooter}{%
  \fancyhf{}
  
  \fancyfoot[L]{\textit{Please cite this article as: Basha, S.H.S., Vinakota, S.K., Dubey, S.R. et al. AutoFCL: automatically tuning fully connected layers for handling small dataset. Neural Comput \& Applic (2021). https://doi.org/10.1007/s00521-020-05549-4. }}
}

\begin{document}
\maketitle
\thispagestyle{specialfooter}
\pagestyle{empty}

\begin{abstract}
Deep Convolutional Neural Networks (CNN) have evolved as popular machine learning models for image classification during the past few years, due to their ability to learn the problem-specific features directly from the input images. The success of deep learning models solicits architecture engineering rather than hand-engineering the features. However, designing state-of-the-art CNN for a given task remains a non-trivial and challenging task, especially when training data size is less. To address this phenomena, transfer learning has been used as a popularly adopted technique. While transferring the learned knowledge from one task to another, fine-tuning with the target-dependent Fully Connected (FC) layers generally produces better results over the target task. In this paper, the proposed AutoFCL model attempts to learn the structure of FC layers of a CNN automatically using Bayesian optimization.
To evaluate the performance of the proposed AutoFCL, we utilize five pre-trained CNN models such as VGG-16, ResNet, DenseNet, MobileNet, and NASNetMobile. The experiments are conducted on three benchmark datasets, namely CalTech-101, Oxford-102 Flowers, and UC Merced Land Use datasets. Fine-tuning the newly learned (target-dependent) FC layers leads to state-of-the-art performance, according to the experiments carried out in this research. The proposed AutoFCL method outperforms the existing methods over CalTech-101 and Oxford-102 Flowers datasets by achieving the accuracy of $94.38\%$ and $98.89\%$, respectively. However, our method achieves comparable performance on the UC Merced Land Use dataset with $96.83\%$ accuracy. The source codes of this research are available at \textcolor{blue}{\url{https://github.com/shabbeersh/AutoFCL}}.

\end{abstract}

\section{Introduction}
\label{sec1}
Deep Convolutional Neural Networks (CNN) based features have outperformed the hand-designed features in most of the computer vision problems such as object recognition \cite{krizhevsky2012imagenet,szegedy2015going}, speech recognition \cite{hinton2012deep}, medical applications \cite{wang2018deep}, and many more. Although several complicated research problems have been solved by deep learning models, generally, the performance of these models relies on hard-to-tune hyperparameters. Finding the best configuration for the hyperparameters such as the number of layers, convolution filter dimensions, number of filters in a convolution layer, and many more to build a CNN architecture suitable for a given task, is the most demanding research theme in the area of Automated Machine Learning (AutoML) \cite{zoph2018learning,liu2018progressive}. Based on the previous studies reported in the literature, learning a suitable architecture for a given task is termed as Neural Architecture Search (NAS) \cite{elsken2018neural}. Reinforcement Learning (RL) methods have been widely employed to find the suitable CNN architecture for given task \cite{jaafra2019reinforcement}. However, these methods are focused to find the structure of CNN from scratch which requires hundreds of GPU hours. We propose a method called AutoFCL to automatically tune the structure of the Fully Connected (FC) layers with respect to the target dataset while transferring the knowledge from source task to the target task.

Typically, every CNN contains one or more FC layers based on the depth of the architecture \cite{basha2019impact}. For instance, the popular CNN models proposed to train over large-scale ImageNet dataset \cite{deng2009imagenet} have the following number of FC layers.
\begin{itemize}
    \item AlexNet \cite{krizhevsky2012imagenet}, ZFNet \cite{zeiler2014visualizing}, and VGG-16 \cite{simonyan2014very} have $3$ dense (FC) layers. Note that these models contain  $5$, $5$, and $13$ convolution layers, respectively.
    \item GoogLeNet \cite{szegedy2015going}, ResNet \cite{he2016deep}, DenseNet \cite{huang2017densely}, NASNet \cite{zoph2018learning}, and other modern deep neural networks have a single FC layer which is responsible for generating the class scores.
\end{itemize}
 
 \begin{figure}[t]
\centering
\includegraphics[width=9 cm , height= 6 cm]{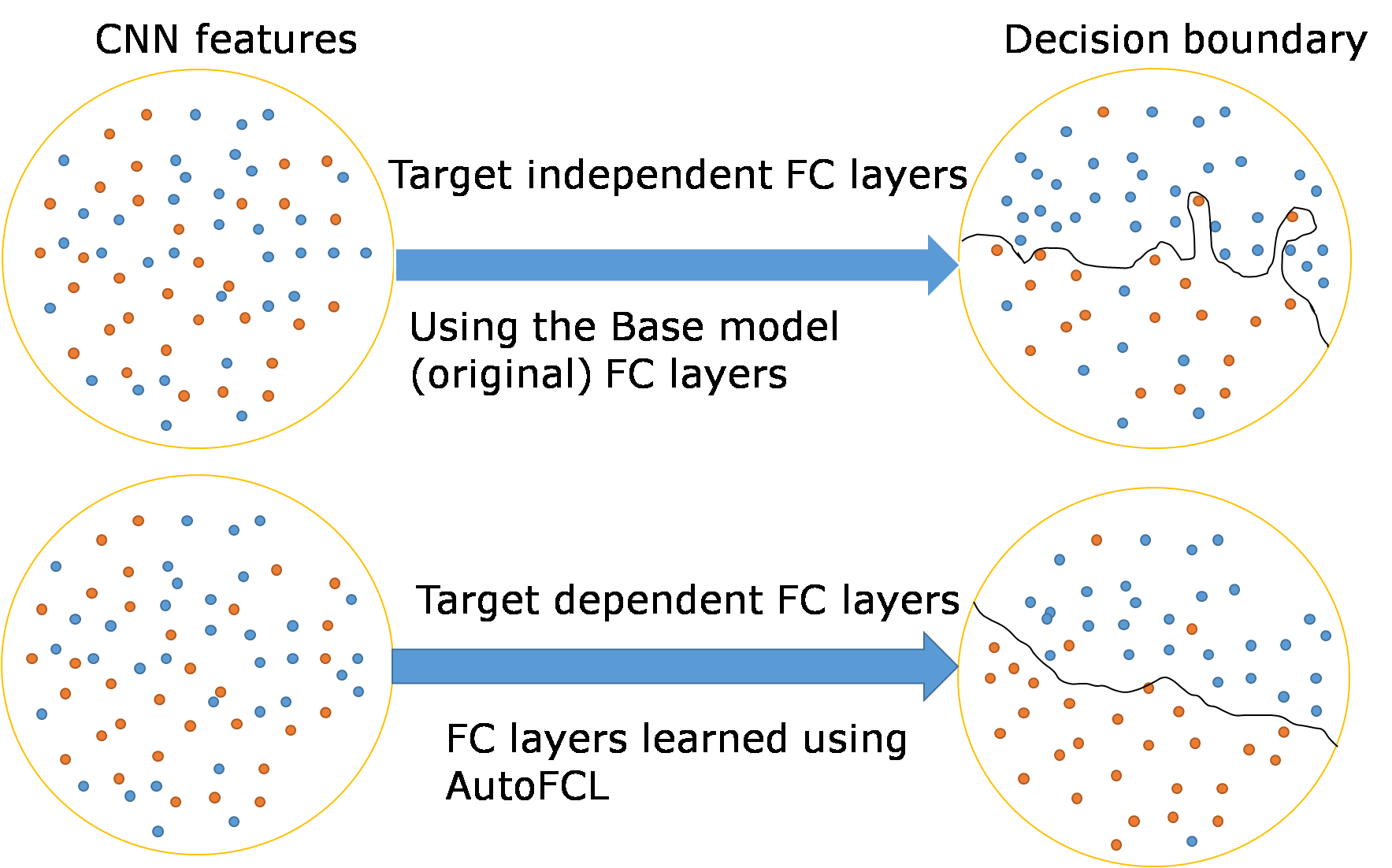} 
\caption{While transferring the knowledge learned from source task to the target task, learning the optimal structure of FC layers with the knowledge of target dataset and fine-tuning the learned FC layers leads to better performance.}
\label{fig:motivation}
\end{figure}
 
The CNN models  \cite{krizhevsky2012imagenet,zeiler2014visualizing,simonyan2014very} introduced in the initial years (during the years from $2012$ to $2014$) have a huge number of trainable parameters in FC layers. Whereas the recent models \cite{he2016deep,huang2017densely,zoph2018learning} are generally deeper, and hence, have a single FC layer which is responsible for generating the class scores. The state-of-the-art CNN architectures proposed for ImageNet dataset are shown in Table \ref{trainable_parameters_table}. This table summarizes the total number of trainable parameters and also the trainable parameters correspond to FC layers. It is evident from \mbox{Table \ref{trainable_parameters_table}} that as the depth of CNN increases, both the number of dense layers and the parameters in dense layers gradually decrease.

A large number of parameters involved in the FC layers of a CNN increases the possibility of overfitting. Xu \etal \cite{xu2019overfitting} shown that removing the connections among FC layers having less weight magnitude (SparseConnect) leads to better performance. Basha \etal \cite{basha2019impact} performed a study to observe the necessity of FC layers given the depth and width of both datasets and CNN architectures. To find the best set of hyperparameters of an Artificial Neural Network (ANN), Mendoza \etal \cite{mendoza2016towards} proposed an automated mechanism to tune the ANN using Sequential Model-based Algorithm Configuration (SMAC).
 
 \begin{table*}
\caption{The state-of-the-art deep neural networks proposed for the ImageNet dataset, the total number of trainable parameters and the number of parameters belong to FC layers are shown.}
\label{trainable_parameters_table}
 \centering
\begin{tabular}{|l|l|l|l|}
\hline
S.No. & CNN Model                                                 & \begin{tabular}[c]{@{}l@{}}Total \#trainable parameters\\ (in Millions)\end{tabular} & \begin{tabular}[c]{@{}l@{}}\#parameters in FC layers \\ (in Millions)\end{tabular} \\ \hline
1    & AlexNet \cite{krizhevsky2012imagenet}    & 60 M                                                                                 & 58 M                                                                               \\ 
2    & ZFNet \cite{zeiler2014visualizing}       &                                                                                     62.3 M &       58.6 M                                                                             \\ 
3    & VGG16 \cite{simonyan2014very}            & 138.3 M                                                                              & 123.6 M                                                                            \\ 
4    & VGG19 \cite{simonyan2014very}            & 143.6 M                                                                              & 123.6 M                                                                            \\ 
5    & InceptionV3 \cite{szegedy2016rethinking} & 23.8 M                                                                               & 2 M                                                                                \\ 
6    & ResNet50 \cite{he2016deep}               & 25.5 M                                                                               & 2 M                                                                                \\ 
7    & MobileNet \cite{howard2017mobilenets}    & 4 M                                                                                  & 1 M                                                                                \\ 
8    & DenseNet201 \cite{huang2017densely}      & 20 M                                                                                 & 1.9 M                                                                              \\ 
9    & NASNetLarge \cite{zoph2018learning}      & 88 M                                                                                 & 4 M                                                                                \\ 
10   & NASNetMobile \cite{zoph2018learning}     & 5 M                                                                                  & 1 M         \\ \hline  
\end{tabular}

\end{table*}
 
CNNs are used in a wide range of applications in recent years. However, their performance is poor if the amount of training data is very limited. Transfer Learning is a way to reduce the need for more training data and huge computational resources by reusing the knowledge learned from the source task.
A common approach for classifying such limited images is re-using the pre-trained models to fine-tune over other datasets \cite{ng2015deep}. However, while transferring the learned knowledge from one task to another, fine-tuning the original FC layers' structure may not perform well over the target dataset because the FC layers are designed for the source task.

Fig. \ref{fig:motivation} illustrates the motivation behind learning the target-dependent fully connected layer's structure to obtain better performance over the target task. While transferring the learned knowledge from source task to target task, the efficacy (capacity) of the CNN increases for the target task, which may result in overfitting. The extracted features from convolutional layers (shown in the left side) are mapped into more linearly separable feature space (shown on the right side) by FC layers. Moreover, we believe that learning the FC layers' structure with the knowledge from the target dataset may lead to better linearly separable feature space which results in better performance over the target dataset. 
In this work, we propose a novel framework for automatically learning the target-dependent fully connected layers structure in the context of transfer learning. We use Bayesian optimization \cite{frazier2018tutorial} for optimizing the hyperparameters involved in forming the FC layers while transferring the knowledge from one task to another.

 

\section{Related Works}
\label{sec_related_works}
Due to the dense connectivity among the FC layers, the deep CNNs contain an enormous amount of trainable parameters. For example, the first ImageNet Large Scale Visual Recognition Competition (ILSVRC)-2012 \cite{ILSVRC15} winning CNN model called AlexNet \cite{krizhevsky2012imagenet} contains a total of $60$ million trainable parameters, among which $58$ million parameters belong to the FC layers. Likewise, VGG-16 \cite{simonyan2014very}, a $16$ layer deep CNN comprises $138$ million trainable parameters, among which $123$ million parameters correspond to FC layers. In practice, the over-parameterization leads to overfitting the CNN. Xu \etal \cite{xu2019overfitting} proposed SparseConnect model to reduce the overfitting effect by removing the connections with smaller weight values.


Transfer learning is a widely adopted technique to obtain a reasonable performance with limited data and less computational resources. Li \etal \cite{li2020transfer} analyzed various approaches for transferring the knowledge learned in different scenarios.
 Fine-tuning the deep CNNs with limited training data often leads to overfitting the CNN model \cite{hu2017discriminative}. Han \etal \cite{han2018new} introduced a two-phase strategy by combining transfer learning with web data augmentation to reduce the amount of over-fitting. They also tuned the hyperparameters such as learning rate, type of optimizer (Adagrad \cite{duchi2011adaptive}, Adam \cite{kingma2014adam}, etc.) and many more using Bayesian Optimization. 

Mendoza \etal \cite{mendoza2016towards} proposed Auto-Net, which automatically tunes an artificial neural network without any human intervention. To learn a distinct set of hyperparameters automatically, they used the Sequential Model-based Algorithm Configuration (SMAC). The hyperparameters such as the number of FC layers, number of neurons in each FC layer, batch size, learning rate, and so on are tuned automatically. Motivated by this work, we propose a framework to automatically learn the structure of FC layers concerning the target dataset for better transfer learning.

Many researchers have employed Bayesian Optimization \cite{frazier2018tutorial} to learn the entire CNN architecture automatically. Wistuba \etal \cite{wistuba2017bayesian} combined Bayesian Optimization with Incremental Evaluation to find the optimal neural network architecture. However, they limited the depth of the CNN to 5 layers due to the limited computational resources. Jin \etal \cite{ji2019novel} proposed a network morphism mechanism for neural architecture search using Bayesian Optimization.
Liu \etal \cite{liu2018progressive} proposed a method to build the CNN architecture progressively using the Sequential Model-Based Optimization (SMBO) based algorithm. However, these methods require a considerable amount of computational resources and search time. Recently, Gupta \etal \cite{gupta2017insights} employed Bayesian Optimization to conduct a study for efficient transfer optimization. 

Transfer Learning allows the pre-trained networks to adopt for the new tasks \cite{yosinski2014transferable}. Many researchers utilized the advantage of transfer learning for various applications \cite{ng2015deep,xie2016transfer}. Ji \etal \cite{ji2019novel} proposed a framework called Double Reweighting Multi-source Transfer Learning (DRMTL) to utilize the decision knowledge from multiple sources to perform well over the target domain. Generally, after adaptation, the efficacy (capacity) of the CNN increases for the target task. Molchanov \etal \cite{molchanov2016pruning} proposed a framework for iteratively pruning the parameters of a CNN to reduce its capacity for the target task. With regard to our knowledge, no effort has been made in the literature to learn the structure of FC layers automatically for better transfer learning. Neural Architecture Search algorithms consume thousands of GPU hours \cite{zoph2018learning} to find better performing architectures. So, we made this attempt in the context of transfer learning to reduce the architecture search time.

Basha \etal \cite{basha2019impact} analyzed the necessity of FC layers based on the depth of a CNN. However, to conduct this study they performed experiments by adding new FC layers manually before the output FC layer. Moreover, the hyperparameters involved in FC layers like the number of neurons in every FC layer, the dropout factor, type of activation, and so on are chosen manually. In this paper, we attempt to learn the target-dependent FC layers' structure automatically for better transfer learning.   


In brief, our contributions in this work are as follows,
\begin{itemize}
    \item We propose a novel method to automatically learn the target-dependent FC layers structure using Bayesian Optimization.
    \item By conducting experiments on three benchmark datasets, we discover the suitable (target-dependent) FC layers structure specific to the datasets.
    \item The performance of the proposed method is also compared with state-of-the-art transfer learning and non-transfer learning-based methods.
    \item To compare the results obtained using Bayesian Optimization, we employed the random search to find the best set of hyperparameters involved in FC layers.
\end{itemize}

\section{Proposed AutoFCL Model}
\label{proposed_method}
  We formulate the task of learning the structure of fully connected layers as a black-box optimization problem. Let $f$ is an objective function whose objective is to find $x_*$, which is represented as
\begin{equation}
    x_* = \argmax_{x \in \mathcal{H}} f(x)
\end{equation}
where $x \in \mathbb{R}^d$ is the input, usually $d \leq 20$ \cite{frazier2018tutorial}, $\mathcal{H}$ is the hyperparameter space as depicted in Table \ref{hyper_space_table}, and $f$ is a continuous function. Finding the value of function $f$ at $x$ requires training (fine-tuning) the learned FC layers (explored during the architecture search) of a pre-trained CNN (B) on training data (Train$_{Data}$) and evaluating its performance on the held-out (validation) data Val$_{Data}$.  

The $\mathit{x_*}$ is a CNN with an optimal FC layer's structure learned using the Bayesian Optimization for efficient transfer learning. Therefore, the CNN architecture $\mathit{x_*}$ is responsible for maximizing the performance on the Val$_{Data}$. The proposed AutoFCL method is outlined in Algorithm \ref{Auto_FCL_algorithm}. Given the base CNN model (B), hyperparameters search space (Param\_space), Train$_{Data}$, Val$_{Data}$, and the number of epochs (E) to train each proxy CNN as an input, the proposed method learns the most suitable structure of FC (dense) layers using Bayesian Optimization \cite{frazier2018tutorial}.



\begin{algorithm*}

\caption{AutoFCL: A Bayesian Search method for automatically learning the structure of FC layers} 
\textbf{Inputs:} B (Base Model),  Param\_space, Train$_{Data}$, Val$_{Data}$, E (num epochs). \\
\textbf{Output:} A CNN with target-dependent FC layers structure.

\label{Auto_FCL_algorithm}
\begin{algorithmic}[1]
 \Procedure{AutoFCL}{} 
 \State Place a Gaussian Process (GP) prior on the objective f
 \While{$t \in 1,2, ...n_{0}$} \Comment{Observe the value of f at initial $n_{0}$ points}
 
 \State $M_{t}\gets build\_CNN(B, Param\_space)$ 
 \Comment{sample the initial CNN randomly} 
 
 \State $T_{t}\gets Train\_CNN(M_{t}, Train_{Data}, E)$ 
 
  \State $V_{t}\gets Validate\_CNN(T_{t}, Valid_{Data})$
 
 \EndWhile
 
$n = n_{0}$ 
 
\While{$t \in n+1, .., N$}   

\State Update the posterior distribution on f using the prior \Comment{Using Eq. \ref{conditional_bayes} }
\State Choose the next sample $x_t$ that maximizes the acquisition function value

 
 \State Observe $y_t = f(x_t)$
\EndWhile
  
\State \textbf{return} $x_t$\Comment{return a point with best FC layer structure}

\EndProcedure

\end{algorithmic}
\end{algorithm*}

The Bayesian Optimization (Bayes Opt) is the most popular method used for finding the best set of hyperparameters involved in deep neural networks \cite{snoek2012practical}. Bayes Opt builds a surrogate model to approximate the objective function using Gaussian Process (GP) regression \cite{williams2006gaussian}. Algorithm \ref{Auto_FCL_algorithm} observes the value of $f$ without noise for initial $n_0$ points which are chosen uniformly random ($n_0$ is 20 in our experimental settings). After observing the objective at initial $n_0$ points, we can infer the objective value at a new point $x_{new}$ using Bayes rule \cite{rasmussen2003gaussian} as follows, 

\begin{equation}
    f(x_{new})|f(x_{1:n_{0}}) \sim Normal(\mu_{n_0}(x_{new}),\sigma^2_{n_{0}}(x_{new}))
    \label{conditional_bayes}
\end{equation}

The $\mu_{n_{0}}(x_{new})$ and $\sigma^2_{n_{0}}(x_{new})$ are computed as follows,

\begin{dmath}
\mu_{n_{0}(x_{new})}=\sum_0(x_{new}:x_{1:n_{0}})\sum_0(x_{1:n_{0}},x_{1:n_{0}})^{-1}(f(x_{1:n_{0}})
- \mu_{0}(x_{1:n_{0}})) + \mu_0({x_{new}})
\end{dmath}


\begin{dmath}
\sigma^{2}_{n_{0}}(x_{new}) = \sum_0(x_{new},x_{new})-\sum_0(x_{new}, x_{1:n_{0}})
\sum_0(x_{1:n_{0}},x_{1:n_{0}})^{-1}\sum_0(x_{1:n_{0}}, x_{new})
\end{dmath}

The probability distribution given in Eq. \ref{conditional_bayes} is called posterior probability distribution. In the above equations, $\mu_0$ , $\sum_{0}$ are mean function and covariance functions, respectively.    

The optimal configuration of FC layers is one among the previously evaluated points (initial $n_{0}$ points) with the maximum f value $(f(x^{+}))$. Now, if we want to evaluate the value of objective 'f' at a new point $x_{new}$, which is observed as $f(x_{new})$. After evaluating the value of $f$ at iteration $n_{0}+1$, the optimal f value will be either $f(x_{new})$ (if $f(x_{new}) \ge f(x^{+})$) or $f(x^{+})$ (if $f(x^{+}) \ge f(x_{new}))$. The improvement or gain in the objective f is $f(x_{new}) - f(x^{+})$ if its value is positive, or 0 otherwise.

However, the f($x_{new}$) value is unknown until observing its value at $x_{new}$ which is typically expensive. Instead of evaluating f at $x_{new}$, we can compute the Expected Improvement (EI) and choose the $x_{new}$ that maximizes the value of EI. Expected Improvement \cite{jones1998efficient} is the most commonly used acquisition function for guiding the search process by proposing the next point to sample. 

For a specified input $x_{new}$, EI can be represented as,
\begin{equation}
    EI(x) = \mathop{\mathbb{E}}[max(f(x_{new})-f(x^+), 0)]
\end{equation}

where $f(x^{+})$ is the maximum validation accuracy obtained so far and $x^{+}$ is the FC layer's structure for which best validation accuracy is obtained. Formally, $x^{+}$ can be represented as,
\begin{equation}
x^{+} = \argmax_{x_{i}\in x_{1:n_{0}}} f(x_{i})    
\end{equation}
which utilizes the information about the models that were already explored and finds the next point that maximizes the expected improvement. After observing the objective at each point, we update the posterior distribution using the Eq. \ref{conditional_bayes}. 

\section{Hyperparameter Search Space}
\label{sec_hyper_space}

This section provides a detailed discussion about the search space used for finding the target-dependent FC layer's structure for efficient transfer learning. A single fully connected layer of a CNN involves various hyperparameters. To mention a few, the number of neurons, dropout rate, and many more. The proposed AutoFCL aims to learn the suitable structure for the FC layers, which includes the number of FC layers, dropout rate, type of activation, and the number of neurons in each FC layer to obtain the better performance over the target dataset. Table \ref{hyper_space_table} shows the hyperparameter search space considered in our experimental settings.

\begin{table*}
\begin{center}
\caption{Hyperparameter search space considered in this paper, which includes both network hyperparameters such as the number of fully connected layers and per-layer hyperparameters like activation function, dropout factor, and the number of neurons are presented in this table.}
\begin{tabular}{|c|l|c|c|}
\hline
 & Name  & Values  & Type\\
\hline\hline
Network hyperparameters & number of FC layers & \{1,2,3\}  &  integer \\ \hline
 Hyperparameters per single FC layer & activation function & \{ReLu, Tanh, Sigmoid\} &   categorical \\
 & dropout rate & \{0.0, 0.1, 0.2, 0.3, 0.4, 0.5\} &  float \\
 & number of neurons & \{64, 128, 256, 512, 1024\} &  integer \\
\hline
\end{tabular}
\label{hyper_space_table}
\end{center}
\end{table*}

As most of the CNN architectures available in the literature have a maximum of $3$ FC layers \cite{krizhevsky2012imagenet, simonyan2014very} including the output layer. Therefore, we consider the search space for the number of FC layers in the range [1,3] (i.e., 1, 2, and 3). The other important hyperparameter is the number of neurons required in each FC layer, for which the proposed method finds the best set of configuration within the range $[64, 1024]$ in powers of 2 (\{64, 128, 256, 512, 1024\}). Besides these hyperparameters, we consider activation function as another hyperparameter. Three popular non-linear activations Sigmoid, Tanh, and ReLu are utilized for the same. To reduce the over-fitting caused due to a large number of trainable parameters in FC layers, dropout \cite{krizhevsky2012imagenet} is widely adopted in deep learning. We consider dropout as another hyperparameter to learn, the value of which is learned in the range [0, 0.5] with an offset 0.1 i.e., the proposed AutoFCL finds the suitable dropout factor within the values \{0.0, 0.1, 0.2, 0.3, 0.4, 0.5\}.

\section{Experimental Settings}
\label{exps_results}
In this section, we brief the training details, CNN architectures utilized to learn the structure of FC layers and the datasets used to evaluate the performance of the developed image classification models in the context of transfer learning.

\subsection{Training Details}
\label{training_details}
\textbf{Training Proxy CNNs:} The CNN architectures generated in the search process of Bayesian Optimization (also called proxy CNNs) are trained using AdaGrad optimizer \cite{duchi2011adaptive}. The initial value of the learning rate is set to $0.01$ and its value is reduced by a factor of $\sqrt{0.1}$ for every $5$ epochs if there is no reduction in the validation loss. Since training the CNNs is a time-consuming task, we train each proxy CNN for $20$ epochs as in \cite{liu2018progressive}. Batch Normalization \cite{ioffe2015batch} is used after employing dropout. The suitable dropout rate is learned using Bayesian Optimization. The parameters (weights) corresponding to the FC layers are initialized using He normal initialization \cite{he2015delving}.  

\subsection{CNN Architectures used for Fine-Tuning}
\label{CNNs_finetuning}
To learn the target-dependent FC layers structure automatically, we use two kinds of CNN architectures which include i) chain structured (plain) CNNs like VGG-16 \cite{simonyan2014very} and ii) CNNs involving skip connections like ResNet \cite{he2016deep}, DenseNet \cite{huang2017densely}, and many more.

We conduct the experiments using the popular CNN models that are trained over ImageNet dataset such as VGG-16 \cite{simonyan2014very}, ResNet \cite{he2016deep}, DenseNet \cite{huang2017densely}, MobileNet \cite{howard2017mobilenets}, and NASNet-Mobile \cite{zoph2018learning}. In this article, we are interested in finding the optimal structure of fully connected layers for efficient transfer learning. To achieve this objective, the parameters (weights) involved in convolution layers of the above CNNs trained over ImageNet dataset \cite{deng2009imagenet} are frozen. In other words, the convolution layers of the above CNNs use the pre-trained weights of ImageNet dataset. The parameters involved in newly added FC layers are learned using the back-propagation algorithm \cite{kelley1960gradient}. The structure of the FC layers is tuned automatically using Algorithm \ref{Auto_FCL_algorithm}.

\begin{figure*}[!t]
\centering
    \subfigure[]{\includegraphics[width=0.32\textwidth]{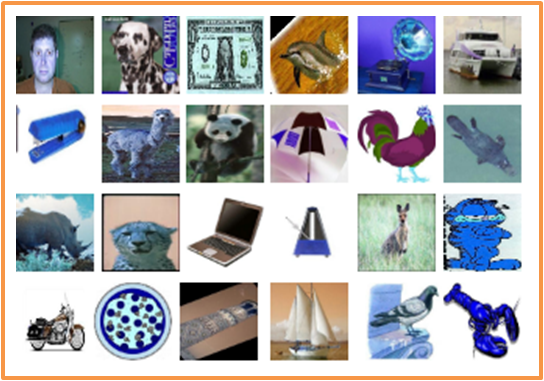}} \label{caltech101_dataset}
    \subfigure[]{\includegraphics[width=0.33\textwidth]{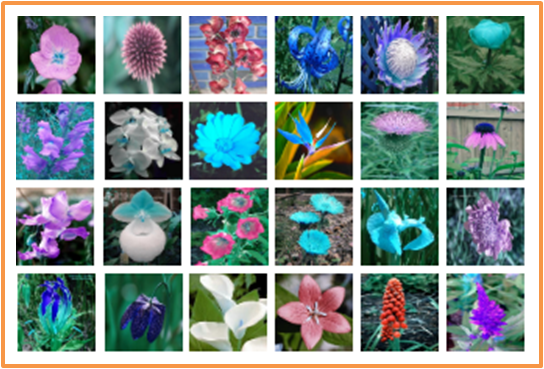}}\label{flowers_dataset}
   \subfigure[]{\includegraphics[width=0.33\textwidth]{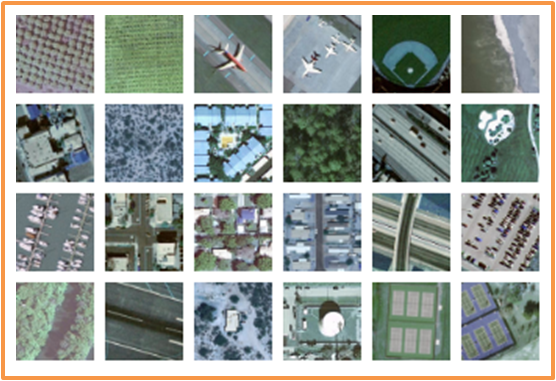}}\label{ucm_dataset}
   
 \caption{(a) A few set of images belong to CalTech-101 \cite{fei2006one}. (b) A few random images from Oxford-102 Flowers \cite{Nilsback08}. (c) Some example images from UC Merced Land Use dataset \cite{yang2010bag}.  }
\label{image_samples}
\end{figure*}

\begin{table*}[!t]
\begin{center}
\caption{The best set of FC layers' hyperparameters learned for CalTech-101 dataset using the Bayesian search and random search techniques. The optimal structure of fully connected layers (excluding the output FC layer) for popular CNNs such as VGG-16, ResNet-50, MobileNet, DenseNet-121, and NASNet-Mobile is presented.}
\label{tab:results_comparison_hyperparameters}

\begin{tabular}{|c|c|c|c|c|c|c|c|}
\hline
S.No               & Model                          & Search Method & \#FC layers & Activation & \#neurons & dropout rate & validation accuracy \\ \hline
\multirow{2}{*}{1} & \multirow{2}{*}{VGG-16}        & Bayesian      & 1           & ReLu       & 256       & 0            & 92.72               \\ \cline{3-8} 
                   &                                & random        & 1           & ReLu       & 512       & 0.3           & 92.34               \\ \hline
\multirow{2}{*}{2} & \multirow{2}{*}{ResNet}        & Bayesian      & 0           & -          & -         & -            & 90.15               \\ \cline{3-8} 
                   &                                & random        & 1           & Sigmoid    & 256       & 0.2          & 89.83               \\ \hline
\multirow{2}{*}{3} & \multirow{2}{*}{MobileNet}     & Bayesian      & 1           & ReLu       & 1024      & 0.3            & 92.50               \\ \cline{3-8} 
                   &                                & random        & 1           & ReLu       & 256       & 0            & 88.73               \\ \hline
\multirow{2}{*}{4} & \multirow{2}{*}{DenseNet}      & Bayesian      & 1           & Sigmoid    & 1024      & 0.3          & 90.21               \\ \cline{3-8} 
                   &                                & random        & 1           & ReLu       & 1024      & 0            & 88.79               \\ \hline
\multirow{2}{*}{5} & \multirow{2}{*}{NASNet-Mobile} & Bayesian      & 1           & ReLu       & 1024      & 0.1          & 88.51               \\ \cline{3-8} 
                   &                                & random        & 1           & Sigmoid    & 256       & 0            & 86.65               \\ \hline
\end{tabular}

\end{center}
\end{table*}

\begin{table*}[!t]
\centering
\caption{Results comparison between the proposed AutoFCL and the state-of-the-art methods over CalTech-101, Oxford-102 Flowers, and UC Merced Land Use datasets. The state-of-the-art including both transfer learning-based and non-transfer learning-based methods are listed in this table. The rows corresponding to the best and second-best performance over each dataset are highlighted in \textbf{bold} and \textbf{\textit{bold-italic}}, respectively.}

\begin{tabular}{|c|l|l|l|}
\hline
\multicolumn{1}{|l|}{Dataset}        & Method                                                                                    & Accuracy       & Transfer Learning/Non Transfer Learning                              \\ \hline
\multirow{10}{*}{CalTech-101}        &  Lee \etal \cite{lee2009convolutional}                     & $65.4$    & Non Transfer Learning                                                \\ \cline{2-4} 
                                     & Cubuk \etal \cite{cubuk2019autoaugment}                   & 86.9      & Transfer Learning                                                    \\ \cline{2-4} 
                                     &  Sawada \etal \cite{sawada2019improvement}                  & 91.8      & Transfer Learning                                                    \\ \cline{2-4} 
                                     & \textbf{Ours (VGG-16 + AutoFCL) }                                                                 &   \textbf{94.38 $\pm$ 0.005}       &  \textbf{Transfer Learning}            \\ \cline{2-4} 
                                     & Ours (ResNet-50 + AutoFCL)                                                       & \textbf{\textit{$91.13 \pm 0.004$}}           & Transfer Learning           \\ \cline{2-4} 
                                     & \textbf{\textit{Ours (MobileNet + AutoFCL)}}                                                                 & \textbf{\textit{92.07 $\pm$ 0.004 }}          & \textbf{\textit{Transfer Learning}}            \\ \cline{2-4} 
                                     & Ours (DenseNet-121+ AutoFCL)                                                              &  $89.5 \pm 0.005 $         & Transfer Learning          \\ \cline{2-4} 
                                     & Ours (NASNetMobile+ AutoFCL)                                                              &  $87.77 \pm 0.005 $         & Transfer Learning  \\ \hline
\multirow{11}{*}{Oxford-102 Flowers} &  
Huang \etal \cite{huang2018flower}                        & 85.66     & Non Transfer Learning                                                \\ \cline{2-4} 
                                     &  Lv \etal \cite{lv2018metric}                              & $92.00$ & Non Transfer Learning                                                \\ \cline{2-4} 
                                     & Murabito \etal \cite{murabito2018top}                     & $79.4$  & Non Transfer Learning                                                \\ \cline{2-4} 
                                     &  Simon \etal \cite{simon2018whole}                           & 97.1   & Transfer Learning                                                   \\ \cline{2-4} 
                                     &  Karlinsky \etal \cite{karlinsky2019repmet}                & $89$    & Transfer Learning                                                    \\ \cline{2-4} 
                                     & \textbf{Ours (VGG-16 + AutoFCL)}                                                                   &    \textbf{98.83 $\pm$ 0.001}       & \textbf{Transfer Learning}               \\ \cline{2-4} 
                                     & \textbf{\textit{Ours (ResNet-50 + AutoFCL)}}                                                                 & \textbf{\textit{97.21 $\pm$ 0.05 }}            &  \textbf{\textit{Transfer Learning}}            \\ \cline{2-4} 
                                     & Ours (MobileNet + AutoFCL)                                                                & $58.6 \pm 0.04$          & Transfer Learning            \\ \cline{2-4} 
                                     & Ours (DenseNet-121 + AutoFCL)                                                              &  $60.91 \pm 0.03$         & Transfer Learning           \\ \cline{2-4} 
                                     & Ours (NASNetMobile + AutoFCL)                                                              &  $41.3 \pm 0.006 $         & Transfer Learning  \\ \hline

    \multirow{8}{*}{UC Merced Land Use}          
    
     &

Shao \etal \cite{shao2013hierarchical}               & 92.38 & Non Transfer Learning                                                \\ \cline{2-4} 
    &

Yang \etal \cite{yang2016bi}               & 93.67 & Non Transfer Learning                                                \\ \cline{2-4} 
                                     &
                                  \textbf{Akram \etal \cite{akram2018deep}}                  & \textbf{97.6}    & \textbf{Transfer Learning}                                      \\
\cline{2-4} 
&
Wang \etal \cite{wang2019deep}        & 94.81  & Transfer Learning                                                    \\ \cline{2-4} 
                                     & 
                                           \textbf{\textit{Ours (VGG-16 + AutoFCL)}}                                                            &   \textbf{\textit{96.83 $\pm$ 0.006}} & \textbf{\textit{Transfer Learning}}         \\ \cline{2-4} 
                                     & Ours (ResNet-50 + AutoFCL)                                                                & $78 \pm 0.03$          & Transfer Learning            \\ \cline{2-4} 
                                     & Ours (MobileNet + AutoFCL)                                                                & $88 \pm 0.004$          & Transfer Learning         \\ \cline{2-4} 
                                     & Ours (DenseNet-121 + AutoFCL)                                                              &   $80.8 \pm 0.015 $       & Transfer Learning           \\ \cline{2-4} 
                                     & Ours (NASNetMobile + AutoFCL)                                                              &   $72.28 \pm 0.016 $       & Transfer Learning  \\ \hline

\end{tabular}

\label{results_comparison_stateoftheart}
\vspace{-4mm}
\end{table*}

\begin{table*}
\centering
\caption{The optimal structure of FC layers learned for Oxford-102 Flowers dataset using the Bayesian search and random search. The values of various hyperparameters for VGG-16, ResNet-50, MobileNet, DenseNet-121, NASNet-Mobile models are shown in this table.}
\begin{center}
\begin{tabular}{|c|c|c|c|c|c|c|c|}
\hline
S.No               & Model                          & Search Method & \#FC layers & Activation & \#neurons & dropout rate & validation accuracy \\ \hline
\multirow{2}{*}{1} & \multirow{2}{*}{VGG-16}        & Bayesian      & 1           & ReLu       & 256       & 0            & 96.64               \\ \cline{3-8} 
                   &                                & random        & 1           & ReLu       & 64        & 0.1          & 94.33               \\ \hline
\multirow{2}{*}{2} & \multirow{2}{*}{ResNet}        & Bayesian      & 1           & Sigmoid    & 512       & 0.3          & 96.31               \\ \cline{3-8} 
                   &                                & random        & 0           & -          & -         & -            & 91.73               \\ \hline
\multirow{2}{*}{3} & \multirow{2}{*}{MobileNet}     & Bayesian      & 1           & Sigmoid    & 512       & 0.5          & 61.29               \\ \cline{3-8} 
                   &                                & random        & 1           & Sigmoid    & 512       & 0.1            & 55.67               \\ \hline
\multirow{2}{*}{4} & \multirow{2}{*}{DenseNet}      & Bayesian      & 1           & Sigmoid    & 1024      & 0.3            & 68.06               \\ \cline{3-8} 
                   &                                & random        & 1           & ReLu       & 256       & 0            & 55.18               \\ \hline
\multirow{2}{*}{5} & \multirow{2}{*}{NASNet-Mobile} & Bayesian      & 1           & ReLu       & 256       & 0.2           & 40.37               \\ \cline{3-8} 
                   &                                & random        & 1           & ReLu       & 512       & 0.1            & 38.37               \\ \hline
\end{tabular}   

\end{center}
\label{results_comparison_hyperparameters_Oxford}
\end{table*}

\subsubsection{Chain Structured CNNs (Plain CNNs)}
In the initial years of deep learning, the CNN architectures proposed such as LeNet \cite{lecun1998gradient}, AlexNet \cite{krizhevsky2012imagenet}, ZFNet \cite{zeiler2014visualizing}, and VGG-16 \cite{simonyan2014very} have the varying number of trainable layers (convolution, Batch Normalization, and fully connected layers) and involves a different set of hyper-parameters. However, the connectivity among the different layers in these architectures remains the same such that layer $L_{i+1}$ receives the input feature map from layer $L_{i}$. Similarly layer $L_{i+2}$ receives the input from layer $L_{i+1}$ and so on. We consider VGG-16, a $16$ layer chain structured deep CNN to learn the structure of FC layers for efficient transfer learning.

\subsubsection{CNNs involving Skip Connections}
Szegedy \etal \cite{szegedy2015going} introduced a deep CNN named GoogLeNet with a careful handcrafted design which allows increasing the depth of the model. GoogLeNet has a basic building block called Inception block that uses multi-scale filters. 
Later on, the concept of skip connections became very popular after the emergence of ResNet in 2016 \cite{he2016deep}. The skip connections are also used by recent models such as DenseNet \cite{huang2017densely}, etc. Moreover, it also became popular among the CNNs learned using NAS methods such as NASNet \cite{zoph2018learning}, PNAS \cite{liu2018progressive}, etc. A layer in the CNNs involving skip connections receives multiple input feature maps from its previous layers. For example, layer $L_{i+1}$ receives the input from both layers $L_{i}$ and $L_{i-1}$ as in ResNet \cite{he2016deep}; layer $L_{n}$ receives the input feature map from all of its previous layers \{$L_{1}$, $L_{2}$, ... $L_{n-1}$\} as in DenseNet \cite{huang2017densely}. We utilized ResNet-50, MobileNet, DenseNet-121, and NASNet-Mobile CNNs involving skip connections to learn the structure of FC layers.

\subsection{Datasets}
To validate the performance of the proposed method, experiments are conducted on three different kinds of benchmark datasets such as CalTech-101, Oxford-102 Flowers, and UC Merced Land Use.

\subsubsection{CalTech-101 Dataset}
CalTech-101 \cite{fei2006one} dataset consists of images belong to 101 object categories. Each class has the number of images between $40$ and $800$. The most common image categories such as human faces tend to have more images compared to others. The total number of images are $9144$ and each image has a varying spatial dimension.  To conduct the experiments, we utilize $80\%$ of the data for training (i.e., $7315$ images) and the remaining $20\%$ images to validate the performance of the deep neural networks. To fit these images as input to the CNN models, we re-size the image dimension to $224\times224\times3$. A few samples from CalTech-101 dataset are presented in Fig. \ref{image_samples}(a).

\subsubsection{Oxford-102 Flowers Dataset}
Oxford-102 \cite{Nilsback08} dataset comprises images belong to 102 flower categories that are commonly visible in the United Kingdom. This dataset contains $8189$ images such that each class has a varying number of flower images ranging from $40$ to $258$. We utilize $80\%$ of the dataset ($6551$ images) for training the CNNs and remaining $1638$ images for validating the performance of the CNNs. To input the images to the CNN models, the image dimension is re-sized to $224\times224\times3$. Some example images from Oxford-102 Flowers dataset are shown in Fig. \ref{image_samples}(b).    

\subsubsection{UC Merced Land Use Dataset}
UC Merced Land Use dataset \cite{yang2010bag} contains images belonging to $21$ categories of lands. This dataset has a total of $2100$ images with $100$ images in each class. The developed CNN models have trained over $80$ images in each class, and the remaining $20$ images are used to validate the performance of the models. The image dimensions are resized from $256\times256\times3$ to $224\times224\times3$. A few images from the UC Merced Land Use dataset are shown in Fig. \ref{image_samples}(c). Next we present the result obtained by the proposed method when applied on the benchmark datasets.



\section{Results and Discussions}
The NAS based CNNs available in the literature generally learn better performing CNN architectures for popular image classification datasets such as ImageNet \cite{deng2009imagenet}, CIFAR-10 \cite{krizhevsky2009learning} which have large number of training examples. However, the image datasets having less amount of training data are not experienced with the advantage of NAS methods. In this paper, we utilize three benchmark datasets for automatically tuning the FC layers of CNNs for better transfer learning. Due to the above reason, we compare the results obtained by the proposed AutoFCL with such state-of-the-art methods which include both transfer learning and non-transfer learning based methods.

\subsection{CalTech-101 Image Classification Results}
To learn the best set of hyperparameters involved in the FC layers of a CNN, we employ two popularly adopted search methods in the literature of \mbox{Neural Architecture Search (NAS)}. Those two search \mbox{methods} include \mbox{i) Bayesian Optimization} and ii) Random Search. Random search chooses the hyperparameters to explore randomly. In our experimental settings, the number of iterations for random sampling is set to 100. Table \ref{tab:results_comparison_hyperparameters} presents the comparison among the performance of proxy CNN models (fine-tuning the best FC layer structure learned during the search process) found using Bayesian search and Random search over CalTech-101 dataset. Table \ref{tab:results_comparison_hyperparameters} also lists the best possible set of hyperparameter values like the number of FC layers, type of activation, number of neurons in each FC layer, and the dropout factor for each FC layer that are learned during the search process. For example, the best structure of FC layers learned using Bayesian optimization for VGG-16 results in $92.72\%$ validation accuracy. After finding the best set of FC layers' hyperparameters using Algorithm \ref{Auto_FCL_algorithm}, we fine-tune the FC layers of the developed CNN models over the CalTech-101 dataset. The CNN models 
are trained for $200$ epochs using AdaGrad optimizer \cite{duchi2011adaptive}. We consider the values of other hyperparameters such as the learning rate similar to the setting of training the proxy CNNs explored during the search process. Fine-tuning the FC layers (learned using the proposed AutoFCL) results in state-of-the-art accuracy $94.38\%$ on CalTech-101 dataset.

\subsection{Oxford-102 Flowers Image Classification Results}
The optimal FC layers hyperparameters learned for the Oxford-102 Flowers dataset using Bayesian Optimization and random search are shown in Table \ref{results_comparison_hyperparameters_Oxford}. Similar to CalTech-101 dataset, once the search process is completed, the FC layers of the CNN (the best FC layer structure found during the search process) are fine-tuned over the Oxford-102 Flowers dataset for 200 epochs using AdaGrad optimizer \cite{duchi2011adaptive}. The proposed AutoFCL achieves the state-of-the-art accuracy of $98.83\%$ on Oxford-102 Flowers dataset. Table \ref{results_comparison_stateoftheart} summarizes the performance obtained using the various CNN models with the target-dependent FC layer structure. The VGG-16 and ResNet-50 achieve the best and second-best state-of-the-art accuracy, respectively over Oxford-102 Flowers dataset. 


\begin{table*}
\caption{The FC layers' hyperparameters are tuned for UC Merced Land Use dataset automatically using the Bayesian search and random search are presented.}
\begin{center}
\begin{tabular}{|c|c|c|c|c|c|c|c|}
\hline
S.No               & Model                          & Search Method & \#FC layers & Activation & \#neurons & dropout rate & validation accuracy \\ \hline
\multirow{2}{*}{1} & \multirow{2}{*}{VGG-16}        & Bayesian      & 1           & ReLu       & 512       & 00.3         & 96.42               \\ \cline{3-8} 
                   &                                & random        & 1           & ReLu       & 64        & 0.1          & 95.23               \\ \hline
\multirow{2}{*}{2} & \multirow{2}{*}{ResNet}        & Bayesian      & 1           & Tanh       & 1024      & 0.2          & 83.8               \\ \cline{3-8} 
                   &                                & random        & 1           & Tanh       & 1024      & 0.4          & 82.14                \\ \hline
\multirow{2}{*}{3} & \multirow{2}{*}{MobileNet}     & Bayesian      & 1           & Sigmoid    & 1024      & 0.5          & 89.52               \\ \cline{3-8} 
                   &                                & random        & 1           & ReLu       & 1024      & 0.1          & 87.38               \\ \hline
\multirow{2}{*}{4} & \multirow{2}{*}{DenseNet}      & Bayesian      & 1           & Sigmoid    & 1024       & 0.0          & 82.38               \\ \cline{3-8} 
                   &                                & random        & 1           & ReLu       & 128     & 0.2            & 81.42               \\ \hline
\multirow{2}{*}{5} & \multirow{2}{*}{NASNet-Mobile} & Bayesian      & 1           & ReLu       & 128       & 0            & 74.76               \\ \cline{3-8} 
                   &                                & random        & 1           & Sigmoid    & 512       & 0.4          & 73.33               \\ \hline
\end{tabular}   

\end{center}
\label{results_comparison_hyperparameters_UCM}
\end{table*}

\subsection{UC Merced Land Use Image Classification Results}
\begin{figure}[!t]
\centering
\includegraphics[width=0.5\textwidth]{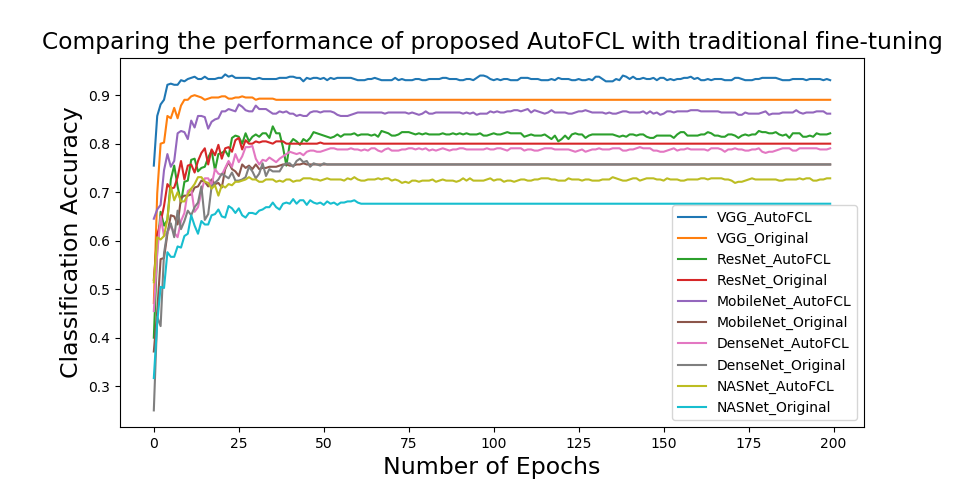} 
\caption{The performance comparison between the proposed AutoFCL and traditional Fine-tuning methods over UC Merced Land Use dataset. Learning the optimal structure of FC layers with the knowledge of the target dataset and fine-tuning the learned FC layers leads to better performance.}
\label{fig:AutoFCL_vs_fine_tuning}
\end{figure}

We consider UC Merced Land Use as another image dataset to learn the best structure of FC layers for efficient transfer learning. The proposed method produces comparable results over UC Merced Land use dataset as presented in Table \ref{results_comparison_stateoftheart}. From Table \ref{results_comparison_stateoftheart} we can observe that fine-tuning the FC layers learned using the proposed AutoFCL for VGG-16 produces $96.83\%$ validation accuracy, which is second best state-of-the-art accuracy. Table \ref{results_comparison_hyperparameters_UCM} lists the best configuration of hyperparameters involved in FC layers found using both Bayesian search and random search. We also compared the performance of the proposed method with fine-tuning original CNN architectures over the target dataset. Fine-tuning the target-dependent FC layer's structure of a CNN over the target dataset results in better performance compared to fine-tuning with the target-independent FC layer's structure. Fig. \ref{fig:AutoFCL_vs_fine_tuning} demonstrates that the proposed AutoFCL outperforms traditional fine-tuning of original FC layers of CNN architectures.   

\section{Conclusion and Future Scope}
\label{conclusion}
We propose AutoFCL, a method to learn the best possible set of hyperparameters belonging to Fully Connected (dense) layers of a CNN for improved transfer learning. Finding the structure of FC layers with the knowledge of target data results in better performance while transferring the knowledge from one task to other. The Bayesian Optimization algorithm is used to explore the search space for the number of FC layers, the number of neurons in each FC layer, activation function and dropout factor. To learn the structure of FC layers, experiments are conducted on benchmark datasets.  The proposed AutoFCL method outperforms the state-of-the-art on most of the datasets. In future, the proposed idea of tuning the pre-trained CNN layers may be extended to tuning the number of Convolution layers of a CNN based on the similarity between the source and target datasets.

\section*{Acknowledgment}

We appreciate NVIDIA Corporation's support with the donation of GeForce Titan XP GPU (Grant number: GPU-900-1G611-2500-000T), which is used for this research.



\bibliographystyle{IEEEtran}  
\bibliography{mybibfile} 
\end{document}